\documentclass[10pt,twocolumn,letterpaper]{article}
\usepackage{epsfig}
\usepackage{bmpsize}

\usepackage{wacv}
\usepackage{times}

\usepackage{amsmath}
\usepackage{amssymb}

\usepackage[utf8]{inputenc} %
\usepackage[T1]{fontenc}    %
\usepackage{hyperref}       %
\usepackage{url}            %
\usepackage{booktabs}       %
\usepackage{amsfonts}       %
\usepackage{nicefrac}       %
\usepackage{microtype}      %
\usepackage{multirow}
\usepackage{xspace}

\usepackage{booktabs} %

\usepackage{mathtools}

\usepackage{times}
\usepackage{amsmath}
\usepackage{amssymb}
\usepackage{bm}
\usepackage{makecell}

\usepackage{microtype}
\usepackage{subfigure}

\usepackage{algorithm}
\usepackage{algorithmic}

\usepackage{float}

\usepackage[toc,page]{appendix}

\usepackage{chngcntr}

\DeclarePairedDelimiter{\ceil}\lceil\rceil

\DeclareMathOperator*{\argmax}{argmax}

\newcommand{\Sec}{Sec.}
\newcommand{\Fig}{Fig.}

\newcommand{\NS}{{NS}\xspace}
\newcommand{\FP}{{FP}\xspace}
\newcommand{\VI}{{VIB}\xspace}
\newcommand{\FG}{{FGA}\xspace}

\wacvfinalcopy %

\def\wacvPaperID{979} %

\ifwacvfinal\fi
\begin{document}

\title{Filter Distillation for Network Compression}

\author{
  Xavier Suau\thanks{Equal contributor} \\
  Apple \\
  \tt\small{xsuaucuadros@apple.com} \\
  \and
  Luca Zappella\footnotemark[1] \\
  Apple \\
  \tt\small{lzappella@apple.com} \\
  \and
  Nicholas Apostoloff \\
  Apple \\
  \tt\small{napostoloff@apple.com}\\
}

\maketitle
\ifwacvfinal\thispagestyle{empty}\fi

\begin{abstract}

In this paper we introduce Principal Filter Analysis (PFA), an easy to use and effective method for neural network compression. PFA exploits the correlation between filter responses within network layers to recommend a smaller network that maintain as much as possible the accuracy of the full model. We propose two algorithms: the first allows users to target compression to specific network property, such as number of trainable variable (footprint), and produces a compressed model that satisfies the requested property while preserving the maximum amount of spectral energy in the responses of each layer, while the second is a {\bf parameter-free} heuristic that selects the compression used at each layer by trying to mimic an ideal set of uncorrelated responses. Since PFA compresses networks based on the correlation of their responses we show in our experiments that it gains the additional flexibility of adapting each architecture to a specific domain while compressing. PFA is evaluated against several architectures and datasets, and shows considerable compression rates without compromising accuracy, e.g., for VGG-16 on CIFAR-10, CIFAR-100 and ImageNet, PFA achieves a compression rate of 8x, 3x, and 1.4x with an accuracy {\bf{gain}} of 0.4\%, 1.4\% points, and 2.4\% respectively. Our tests show that PFA is competitive with state-of-the-art approaches while removing adoption barriers thanks to its practical implementation, intuitive philosophy and ease of use.
\end{abstract}

\section{Introduction}
\label{introduction}

Despite decades of research, the design of deep neural networks (DNNs) is often an empirical process. Practitioners frequently make design choices, such as the number of layers, types of layer, number of filters per layer, etc., based on intuition or brute-force search. Nevertheless, the strong performance of DNNs, together with GPU advances, have led to a growing popularity of these techniques in both academia and industry. Recent studies have unveiled some intrinsic properties of DNNs. For example, there is a consensus that depth can accelerate learning, and that wider layers help with optimization~\cite{Arora:ICML:2018, Livni:NIPS:2014, Nguyen:ICML:2017, Cohen:CORR:2018}. However, in practical applications, the size of these networks is often a limiting factor when deploying on devices with constrained storage, memory, and computation resources.

Another observed DNN property is that the responses of a layer exhibit considerable correlation~\cite{Denil:NIPS:2013}, inspiring the idea of learning decorrelated filters~\cite{Cogswell:ICLR:2016,Rodriguez:ICLR:2017}. ~\cite{Cogswell:ICLR:2016,Rodriguez:ICLR:2017} propose a modified loss function to encourage decorrelation during training and show that accuracy improves with decorrelated filters. However, such algorithms focus on training and do not address network compression. \emph{Our hypothesis is that layers that exhibit high correlation in \textbf{filter responses} could learn equally well using a smaller number of filters.}

Principal Filter Analysis (PFA) draws from the recent findings by letting the user start with an over-parametrized network and then leverages the intra-layer correlation to reduce the network size after training. PFA analyzes a trained network and is agnostic to the training methodology and the loss function. Inference is performed on a dataset, and the correlation within the responses of each layer is used to provide a compression recipe. A new, smaller architecture based on this recipe can then be fine-tuned.

We propose two closed-form algorithms based on spectral energy analysis for suggesting the number of filters to remove in a layer:
\begin{description}
\item[PFA-En] uses Principal Component Analysis (PCA)~\cite{PCA:1933} to allow a user to specify the proportion of the energy in the original response that should be preserved in each layer; since this operation is extremely fast, the user can alternatively provide a network property, such as footprint (or FLOPs), and different energy thresholds can be iteratively tested until the requested property is satisfied.
\item[PFA-KL] is a parameter-free approach that balances the trade-off between compression and accuracy change. PFA-KL uses Kullback-Leibler (KL) divergence \cite{kullback1951} to quantify the divergence between the current set of responses and an ideal set of uncorrelated responses in order to identify the number of redundant filters.
\end{description}

Based on the PFA recommendation, filters that produce maximally correlated responses are removed and the network is fine-tuned. As shown in \Sec~\ref{sec:quantitative}, using popular convolutional networks and datasets such as VGG-16 on CIFAR-10, CIFAR-100 and ImageNet, PFA achieves a compression rate of 8x, 3x, and 1.4x with an accuracy {\bf{gain}} of 0.4\%, 1.4\% points, and 2.4\% respectively. Our tests show that PFA is competitive with state-of-the-art approaches while providing the unique advantage of being {\em practical to implement, intuitive to understand, and easy to use}. %
Since PFA exploits the correlation of the responses, its recommendations become specific for a given domain. This specialization makes PFA suitable also for the task of simultaneous compression and domain adaptation, as shown in ~\Sec~\ref{sec:domain}.

\section{Related work}
\label{soa}

The field of network compression encompasses a wide range of techniques that can be grouped into the following families: quantization, knowledge distillation, tensor factorization and network pruning.

\textbf{Quantization} algorithms compress networks by reducing the number of bits used to represent each weight~\cite{Jiaxiang:CVPR:2016,Han:ICLR:2016,Rastegari:ECCV:2016, He:ECCV:2018, Wu:ICML:2018}. 

\textbf{Knowledge distillation}~\cite{Hinton:NIPS:2014} aim to create a simpler model that mimics the output of a more complex model. Variations on this concept include~\cite{Bucilua:SIGKDD:2006,Ba:NIPS:2014,Romero:ICRL:2015,Chen:ICLR:2016}.

\textbf{Tensor factorization} algorithms exploit the redundancy present in convolution layers by replacing the original tensors with a sequence of smaller or sparser counterparts that produce similar responses~\cite{Denton:NIPS:2014,Lebedev:ICLR:2015,Jaderberg:BMVC:2014,Tai:ICLR:2016,Masana:ICCV:2017,Wen:ICCV:2017,Yu:CVPR:2017,Alvarez:NIPS:2017,Peng:ECCV:2018}. %

\textbf{Network pruning} is a family of techniques that compress networks by iteratively removing connections based on the salience of their weights.  Early work, like Optimal Brain Damage~\cite{Lecun:NIPS:1990} and Optimal Brain Surgeon~\cite{Hassibi:NIPS:1993}, targeted fully connected networks. Recent work can be divided into two sub-families: {\emph {sparse pruning}}~\cite{Han:NIPS:2015,Srinivas:BMVC:2015,Wen:NIPS:2016,Tung:CVPR:2018,Aghasi:NIPS:2017, Carreira:CVPR:2018, Zhang:ECCV:2018, Dai:ICML:2018}, where individual neurons are removed, and {\emph {structured pruning}}~\cite{Li:ICLR:2017,He:ICCV:2017,Luo:ICCV:2017,Molchanov:ICLR:17,Yu:CVPR:2018}, where entire filters are removed.

PFA falls within the family of structured network pruning. Some of these techniques (e.g., \cite{Li:ICLR:2017}) require user defined parameters that are hard to choose and whose effect on the footprint is difficult to predict (see \Sec~\ref{sec:limitations} for a more detailed discussion). Others also require modification of the loss function (e.g., \cite{Liu:ICCV:2017}). In contrast, PFA-En has only one and intuitive parameter, which is the proportion of the response energy to be preserved at each layer, and PFA-KL is parameter-free. Furthermore, instead of learning the saliency of the filters during training by modifying the loss function, PFA estimates it after training without requiring knowledge of the training details. This makes PFA applicable to any trained network, without the need to know its loss function or training regime. 

Within the structured pruning family, there are approaches based on singular value decomposition (SVD)~\cite{Xue:ICASSP:2013, Nakkiran:IS:2015, Prabhavalkar:ICASSP:2016}, where a new set of filters are obtained by projecting the original weights onto a lower dimensional space. Techniques that make compression decisions based on {\em weights}, rather than the {\em responses}, cannot take into account the specificity of the task. PFA differs from these methods because SVD is performed on the responses of the layers rather than on the filter weights, and no projection is done. This is particularly important for domain adaption applications, where a trained network is specialized for a different task. As shown in \Sec~\ref{sec:domain}, PFA derives different architectures from the same initial model when the responses are obtained from different tasks (i.e., datasets). %

Some methods also reason on the layer responses~\cite{Li:ICLR:2017, Masana:ICCV:2017, Peng:ECCV:2018}. These techniques aim to find a smaller set of filters that minimize the reconstruction error of the feature maps or the response output. %
PFA has a different philosophy: it uses the concept of correlation within the responses to identify redundancy within a layer. In practice, this means that PFA can compress all layers simultaneously, while the majority of the techniques that use responses need to operate on one layer at the time.

Finally, PFA is orthogonal to the quantization, tensor factorization and distillation methods, and could be used as a complementary strategy to further compress neural networks.

\section{Principal Filter Analysis}
\label{pfa}

\newcommand{\D}{\mathcal{D}}
\newcommand{\X}{\mathbf{X}}
\newcommand{\x}{\mathbf{x}}
\newcommand{\W}{\mathbf{W}}
\newcommand{\T}{\mathbf{T}}
\newcommand{\act}{\mathbf{a}}
\newcommand{\A}{\mathbf{A}}
\newcommand{\R}{\mathbb{R}}
\newcommand{\p}{p}

\newcommand{\bu}{\bm{u}}
\newcommand{\bd}{\bm{d}}
\newcommand{\br}{\bm{r}}
\newcommand{\corr}{r}

\newcommand{\params}{\bm{\Gamma}}
\newcommand{\param}{\gamma}

\newcommand{\Energy}{\mathcal{E}}
\newcommand{\KL}{\mathrm{KL}}

\newcommand{\foot}{\mathcal{F}}
\newcommand{\footstar}{\foot^{\star}}

\newcommand{\lay}{{[\ell]}}

\newcommand{\bL}{\bm{\lambda}^\lay}
\newcommand{\buL}{\bm{u}^{\lay}}

In this section, the PFA-En and PFA-KL algorithms are described in detail. Both algorithms share the idea of exploiting correlations between responses in convolutional layers and neurons in fully connected layers to obtain a principled recommendation for network compression.%

\subsection{Definitions}
PFA is inherently data driven and thus takes advantage of a dataset $\{\X_i\}\in \R^{M \times I}$ where $\X_i$ is the $i^{th}$ input data sample, $M$ is the number of samples in the dataset, and $I$ is the input dimensionality. Typically, this dataset is the data used to train the network, but it can also be a representative set that covers the distribution of inputs likely to be encountered. Without loss of generality, we assume that the input data are images:$\{\X_i\} \in \R^{M \times H \times W \times C}$, where $H$ is the image height, $W$ is the image width and $C$ is the number channels.

Let $\T_{i}^\lay \in \R^{1 \times H^\lay \times W^\lay \times C^\lay}$ be the output tensor produced by a given layer $\ell$ of a network on the $i^{th}$ input sample. Any operation in the network is considered a layer (e.g., batch normalization, ReLU, etc.). In this work, we analyze the output of convolutional and fully connected layers%
. 

For a convolutional layer $\ell$, let $\W^\lay \in \R^{f_h^\lay \times f_w^\lay \times C^{[\ell-1]} \times C^\lay}$ be the set of $C^\lay$ trainable filters with a kernel of size $f_h^\lay \times f_w^\lay \times C^{[\ell-1]}$. Therefore, we can formally express $\T_{i}^\lay$ produced by the convolutional layer as $ \T_{i}^\lay = \W^\lay * \T_{i}^{[\ell-1]}  \; \text{with} \; \T_{i}^{[0]} = \X_i,$ 
where $ * $ denotes the convolution operator. We omit the bias term to improve readability.

We define the response vector $\act_{i}^\lay \in \R^{C^\lay}$ of a given layer $\ell$ with respect to an input $\X_i$ to be  
the spatially max-pooled and flattened tensor $\T_{i}^\lay$ (i.e., max-pooling over the dimensions $H^\lay$ and $W^\lay$). For fully-connected layers, $\W^\lay \in \R^{C^{[\ell-1]} \times C^\lay}$, with $C^\lay$ being the number of neurons in layer $\ell$. The output tensor is $\T_{i}^\lay = \W^\lay \T_{i}^{[\ell-1]}$, and since no pooling is required, we define the response vector as $\act_{i}^\lay = \T_{i}^\lay \in \R^{C^\lay}$. 

Let $\A^\lay = [\act_{1}^\lay, \ldots, \act_{M}^\lay]^\top \in \R^{M \times C^\lay}$ be the matrix of responses of a generic layer $\ell$ given a dataset with $M$ samples. Given $\A^\lay$ we can compute its covariance matrix $\in \R^{C^\lay \times C^\lay}$, and extract its eigenvalues $\bL \in \R^{C^\lay}$, {\em sorted in descending order and normalized to sum to 1}.\\
\subsection{Compression recipes}
The set $\bL$ provides insight into the correlation of the responses produced by layer $\ell$. {\em The closer $\bL$ is to a uniform distribution, the more decorrelated the response of the filters and the more uniform their contribution to the overall response energy}. Conversely, the closer $\bL$ is to a Dirac $\delta$-distribution, the more correlated the filters. Our hypothesis is that layers that exhibit high correlation could learn equally well using a smaller number of filters.

We present two strategies that use $\bL$ and produce a recipe with the goal of maximizing compression by reducing correlation. Let a recipe $\params = \{\param^\lay\}$, with $\param^\lay \in (0,1]$, be the set of compression factors applied to each of the $L$ layers included in the analysis. For example, $\param^{[3]} = 0.6$ means that we keep $60\%$ of the filters in layer 3. 

Up to now the recipes only indicate \emph{how many} filters each layer should have. Once the correct number of filters has been determined, we continue to choose \emph{which} filters should be kept. We call this {\em filter selection} and we outline it in~\Sec~\ref{sec:filter_selection}.\\
\\
{\noindent{\bf{PFA-En: energy-based recipe.}}} PCA can be used for dimensionality reduction by performing a linear mapping to a lower dimensional space that maximizes the variance of the data in this space. This can be accomplished by extracting the eigenvectors and eigenvalues of the covariance matrix. The original data is then reconstructed using the minimum number of eigenvectors that correspond to the eigenvalues that sum up to the desired energy factor $\tau$. Inspired by this strategy, we propose to keep the minimum set of filters such that a fraction of response energy greater or equal to a user defined energy, $\tau$, is preserved. We define the energy at a given compression ratio for a layer as
$
\Energy(\param^\lay) = \sum_{k=1}^{\ceil*{\param^\lay \cdot C^\lay}}{\lambda_k^\lay},
$
and we propose to re-architect the network according to the following recipe:
\begin{equation}
\label{energy_recipe}
\params^\star_{\Energy}(\tau) = \{\min \param^\lay \} \quad \text{s.t.} \quad \Energy(\param^\lay) \ge \tau, \quad \forall \ell.
\end{equation}
The parameter $\tau$ provides the user with the ability to guide the compression ratio.

PFA-En has the advantage of being tightly connected to well-established dimensionality reduction techniques based on PCA, is simple to implement, and uses a single, highly intuitive parameter. Furthermore, since evaluating the size of a model (or its FLOPs) obtained at different energy thresholds is easy and fast, it is straightforward to replace the parameter $\tau$ with the desired footprint $\foot$ (or FLOPs) after compression by solving iteratively the optimization:
$\params^\star_{foot}(\foot) = \argmax_{\tau} foot(\params^\star_{\Energy}(\tau)) \le \foot$, where $foot(\cdot)$ is a function that returns the footprint of a model given a recipe. Being able to specify a target footprint instead of an energy threshold gives PFA-En an even greater appeal for practical use cases.\\

\noindent{\bf{PFA-KL: KL divergence-based recipe.}}
We propose an alternative formulation to obtain a recipe $\params_\KL^\star$, based on the KL divergence. This formulation is a heuristic that frees PFA from the use of any parameter. As previously mentioned, a set $\bL$ similar to a uniform distribution implies an uncorrelated response of the filters in layer $\ell$. Therefore, the further $\bL$ is from a flat distribution the more layer $\ell$ can be compressed.

Let us define $\buL \in \R^{C^\lay} \sim \cup[1, C^{\lay}]$ as the \textit{desired} uniform (i.e., flat) distribution (no correlation between filters), and $\bd = \text{Dirac}()$ as the \textit{worst case} distribution (all filters are perfectly correlated). We can measure the dissimilarity of the actual set, $\bL$, from the \textit{desired} distribution, $\buL$, as the empirical KL divergence $\KL(\bL,\buL)$. The upper bound of which is given by $u_\KL = \KL(\bd,\buL)$, while the lower bound is 0. Note that the KL divergence is not symmetric, however, since $\bd$ has only one point of support, $u_\KL$ can only be computed in one direction.
Also note that one could replace the KL divergence with any dissimalarity measure between distributions, such as $\chi^2$ or the Wasserstein metric~\cite{Rubner:ICCV:1998}.

Intuitively, when the actual set of eigenvalues is identical to the ideal distribution (i.e., no correlation found) then we would like to preserve all filters. Conversely, when the actual set of eigenvalues is identical to the worst case distribution (i.e., all filters are maximally correlated) then one single filter would be sufficient. The proposed KL divergence-based recipe is a mapping $\psi : [0,u_\KL] \mapsto (0,1]$; a divergence close to the upper bound results in a strong compression and a divergence close to the lower bound results in a milder compression:
\begin{equation}
  \label{kl_recipe}
  \params_\KL^\star =  \Big \{ \psi\big(\KL(\bL,\buL), u_\KL\big) \Big \}, \;\; \forall \ell.
\end{equation}
In this work, we use a simple linear mapping $\psi(x,u_{KL})=1-\nicefrac{x}{u_{KL}}$%
. Other mappings were explored, leading to different degrees of compression; however, we have observed that a linear mapping produces good results that generalize well across networks.%

\subsubsection{Filter selection}\label{sec:filter_selection}
The recipes produced by PFA-En and PFA-KL provide the number of filters, $F^{\lay} = \ceil*{\param^{\lay}\cdot C^{\lay}}$, that should be kept in each layer, but do not indicate which filters should be kept. Once a new compressed architecture is created the question becomes how to initialize it. One option is to initialize it at random. In this case, it does not matter which filters are chosen. An alternative is to select which filters to keep and use their values for initialization, with the intuition (verified in our experiments) that the use of previously trained filters will improve convergence and, for the same given training budget, lead to better accuracy than random initialization.
We do this by removing those filters in each layer that are maximally correlated. For each filter in a given layer we compute the $\ell_1$-norm of the Pearson's correlation coefficients~\cite{Pearson:RSL:1895} with all the other filters	, and remove the filter with the largest norm. If more filters need to be removed, we update the coefficients by removing those that correspond to the previously selected filter, and iterate until the desired number of filters has been removed. In the rare, but theoretically possible, case in which two filters have the same $\ell_1$-norm we choose the one with the highest individual correlation coefficient.

\section{Experiments}
\begin{figure*}[t]
	\centering
	\includegraphics[width=\textwidth]{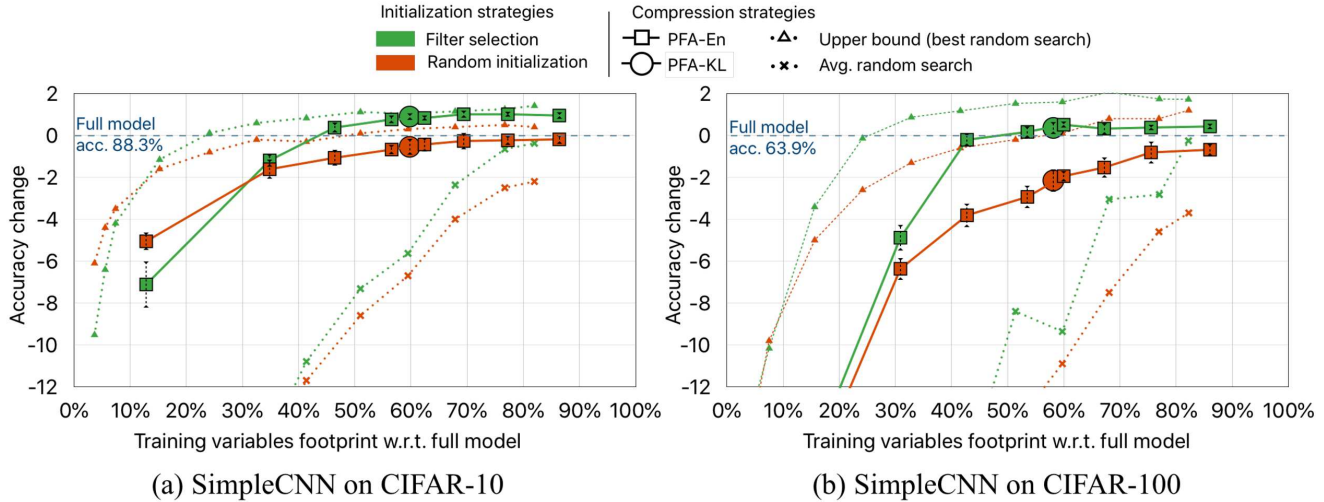}
	\caption	{\label{fig:simplecnn_cifar}  Results of different SimpleCNN compressed networks. 
          Accuracy change in the $y$ axis is reported in percentage points (error bars show the standard deviation of multiple runs). Note how: (1) all PFA solutions lie close to the upper bound while random pruning severely degrades accuracy; (2) in most cases filter selection strategy is better than random initalization.}
\end{figure*}
\label{sec:experiments}

\subsection{Quantitative comparison}\label{sec:quantitative}

To evaluate PFA, we apply it to several architectures and datasets, and we compare results to the state of the art. We compare PFA to another method, the filter pruning approach (\FP)\cite{Li:ICLR:2017}, that like PFA belongs to the family of structured pruning algorithms. We also extend the comparison to other families even if algorithms in those families tend to be more complex and computationally demanding. We compare against sparse pruning algorithms, such as the network slimming approach (\NS)~\cite{Liu:ICCV:2017}, and the variational information bottleneck approach (\VI)~\cite{Dai:ICML:2018}. We also provide a comparison against a tensor facorization method: the filter group approximation approach (\FG)~\cite{Peng:ECCV:2018}.

For the comparison, we focus on the compression ratio and the accuracy change, measured in percentage points (pp), obtained by the compressed architectures. This enables comparing various techniques in the same plot, even if the accuracy of each original architecture is slightly different because of different training strategies used. In App.~\ref{app:details} and \ref{app:training}, we provide the exact accuracy of the full and compressed models, footprint, FLOPs, and all the hyper-parameters used to train the models.\\

\noindent{\bf{Ablation studies and random compression.}}
In this set of experiments we assess the impact of the PFA compression strategies separately from the impact of the initialization of the pruned network, i.e., the filter selection strategy. In addition, we try to understand how PFA compares to randomly compressed networks and how close its solution is to the optimal architecture (empirically defined).

In order to be able to repeat this experiment many times we adopt a small convolutional network that we refer to as SimpleCNN (see App.~\ref{app:simplecnn} for the exact specification of the network and the training hyper-parameters). Results are shown on CIFAR-10 and CIFAR-100~\cite{CIFAR}. The full model is obtained by training using 10 random initializations -- we choose the initialization that leads to the highest test accuracy and perform inference on the training set to obtain the responses ($\A^\lay$) at each layer needed for the PFA analysis. We analyze all layers in parallel (one-shot, as opposed to an iterative approach which would also be applicable to PFA and likely to lead to even better results, but it would not be a fair comparison with all other state-of-art algorithms) to obtain PFA recipes. PFA-En is computed for the following energy values: $\tau \in \{0.8, 0.85, 0.93, 0.95, 0.96, 0.97, 0.98, 0.99\}$, whereas PFA-KL is parameter-free and is computed once per baseline network.

To evaluate different initialization strategies (i.e., random vs filter selection), after creating the compressed architecture according to a PFA recipe we perform two types of fine-tuning. First, we retrain from scratch with 10 different random weight initializations. Second, we retrain 10 times using {\em{filter selection}}, and fine-tune the compressed network starting from the weights of the selected filters. We report the mean and standard deviation of the accuracy of each of these 10 models. It is important to note that the retraining is done without hyper-parameter tuning (we use the same parameters used to train the full model). While this is a sub-optimal strategy, it removes any ambiguity on how well the parameters were tuned for the full model compared to the compressed networks. In practice, one could expect to attain even better results if parameter sweeping was performed on the compressed networks. 

An {\em empirical upper bound} of the accuracy at different footprints is obtained by randomly choosing how many filters to remove at each layer, and by repeating this process a sufficient number of times. The best result at each footprint can be considered an empirical upper bound for that architecture and footprint. On the other hand, the result averaged across all random searches is representative of how easy (or difficult) it is to randomly compress a network without hurting its accuracy. In these experiments we trained {\bf 300} randomly pruned architectures for each footprint. 

Results are reported in \Fig~\ref{fig:simplecnn_cifar}. The first notable remark (for both datasets and initialization strategies) is the considerable gap between the upper bound (up-facing triangles) and the average random search (crosses): this indicates, unsurprisingly, that random pruning is not an effective strategy. The second remark is that there exist smaller architectures derived from the base model that can perform even better than the full model. We attribute this result to the potential of a smaller model to generalize better.

PFA-En (squares) and PFA-KL (circles) are consistently better than the mean random search and very close to the empirical upper bound. The use of filter selection (green) improves in all approaches (even random search) compared to random initialization (brown). Notably, when using PFA with filter selection the accuracy for footprints above 40\% becomes even better than that of the full model.

Interestingly, at the 30\% footprint mark a random initialization for PFA-En appears to be better than the use of filter selection. It is possible that when keeping an extremely small number of filters, the starting point provided by the filter selection becomes a local minimum that is difficult to escape. For thin layers in relatively small architectures (like SimpleCNN), a random initialization may give more room for exploration during the learning phase.

Overall, we have found that the filter selection strategy converges faster during training and performs consistently better across different architectures and datasets, hence, from now on we will only report results using PFA with filter selection.\\

\noindent{\bf{CIFAR-10 and CIFAR-100.}} We repeated the experiments above on known architectures and compared our results with state-of-the-art techniques.

\begin{figure*}[tb]
  \centering
  \subfigure[VGG-16 on CIFAR-10]{\label{fig:vgg16_cifar10} \includegraphics[width=0.49\textwidth]{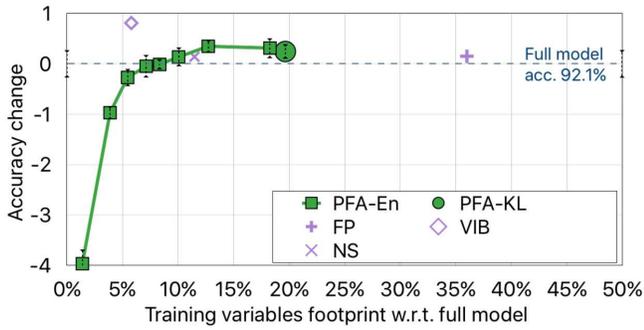}}
  \subfigure[ResNet-56 on CIFAR-10]{\label{fig:resnet56_cifar10} \includegraphics[width=0.49\textwidth]{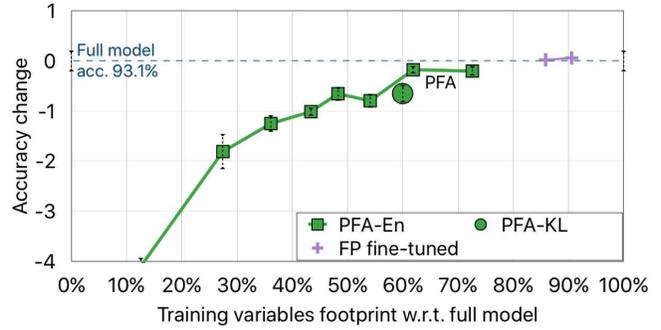}}        
  \subfigure[VGG-16 on CIFAR-100]{\label{fig:vgg16_cifar100} \includegraphics[width=0.49\textwidth]{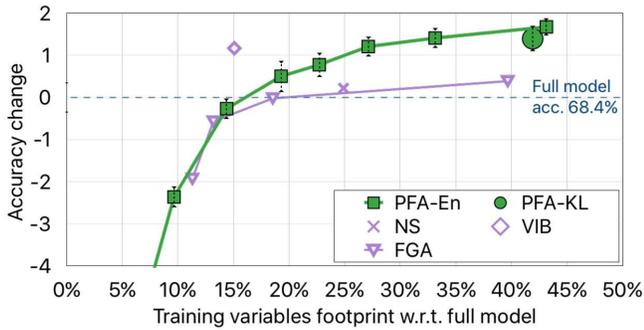}}
  \subfigure[ResNet-56 on CIFAR-100]{\label{fig:resnet56_cifar100} \includegraphics[width=0.49\textwidth]{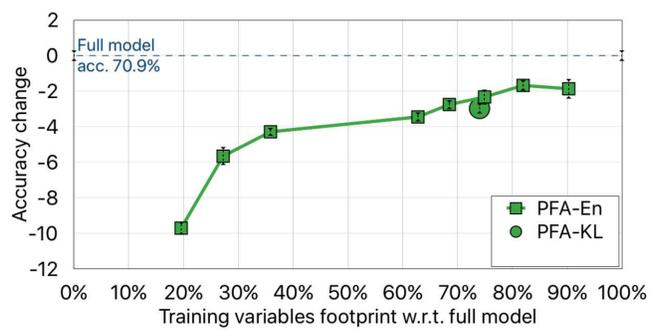}}        
  \caption{Results for VGG-16 and ResNet-56 on CIFAR-10 and CIFAR-100. 
    Accuracy change in the $y$ axis is reported in percentage points. Note how the accuracy obtained by PFA is comparable to the state of the art, and how PFA works across different architectures.}
    \label{fig:cifar10_100}
\end{figure*}

The architectures used are VGG-16~\cite{Simonyan:ICLR:2015} (version proposed by~\cite{Zagoruyko:TORCH:2015} for CIFAR) and ResNet-56~\cite{He:CVPR:2016}, with padding in the skip-connections (refer to App.~\ref{app:training_cifar} for the training hyper-parameters and for a detailed explanation on how we handle skip-connections with padding). We compare the results of PFA with those reported by \FP, \VI\footnotemark, 
\footnotetext{Error of the original full models kindly provided by the authors of \VI.}
\FG\footnotemark, and \NS (after a single iteration for a direct comparison with the other methods).
\footnotetext{Number of trainable variables kindly provided by the authors of \FG.}

As shown in~\Fig~\ref{fig:cifar10_100}, results are consistent irrespective of the architecture and dataset: PFA is comparable to or does better than more complex techniques that require more computational resources such as \NS and \FG. At comparable footprints, \VI achieves a slightly higher accuracy than PFA (around 1 pp), but again at higher computational cost.\\

\noindent{\bf{ImageNet.}}
\begin{figure*}[tb]
  \centering
\subfigure[VGG-16 on ImageNet]{\label{fig:vgg16_imagenet} \includegraphics[width=0.49\textwidth]{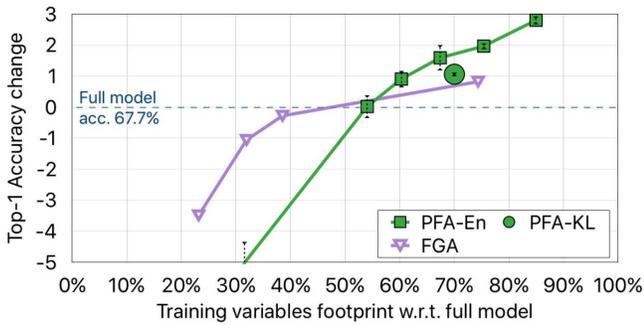}}      
\subfigure[ResNet-34 on ImageNet]{\label{fig:resnet34_imagenet} \includegraphics[width=0.49\textwidth]{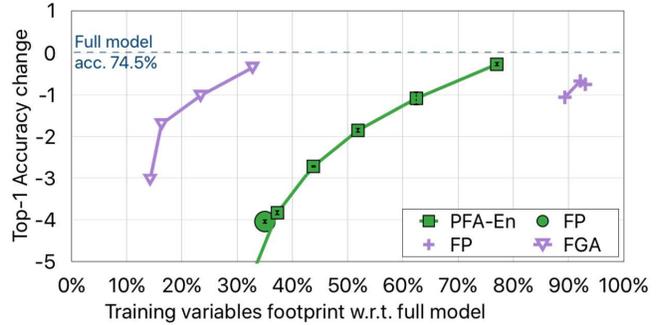}}      
  \caption{Results on ImageNet. 
    Accuracy change in the $y$ axis is reported in percentage points. On VGG-16 PFA is comparable to FGA. On ResNet-34, PFA is better than FP but FGA, in this specific experiment, achieves an even stronger compression.}\label{fig:imagenet_top1}
\end{figure*}
For the experiments on ImageNet~\cite{ImageNet}, we train and compress one baseline for each architecture. We retrain the models obtained by PFA 3 times (using the same hyper parameters used to train the full model) and report the mean and standard deviation of the Top-1 accuracy (we also provide the Top-5 accuracy in the App.~\ref{app:details}).

The architectures used are VGG-16~\cite{Long:CVPR:2015} (fully convolutional) and ResNet-34, with projection in the skip-connections (refer to App.~\ref{app:training_imagenet} for a detailed explanation on how we handle skip-connections with projections).

Results are shown in~\Fig~\ref{fig:imagenet_top1}. On VGG-16, PFA achieves better accuracy than \FG\footnotemark, at comparable model sizes, until a size of 50\% of the full model, after which \FG maintains a better accuracy. On ResNet-34 PFA achieves both a better accuracy and stronger compression than \FP. In this experiment, \FG is extremely efficient and outperforms other techniques, which is different than what we observed in previous experiments, where PFA performed comparably (VGG-16 with ImageNet) or better (VGG-16 with CIFAR-100) than \FG.

\footnotetext{Number of trainable variables kindly provided by the authors of \FG.}

We have shown that PFA works consistently across different architectures and datasets. In general, PFA is comparable to the state of the art even without training hyper-parameters search, and contrary to most state-of-the-art algorithms it does not require tuning of its own parameters.\\

\noindent{\bf{On the complexity and scalability of PFA}}
The complexity of PFA (excluding the inference step), with respect to number of filters and dataset size, is dominated by the the PCA analysis which, for a given layer, is $O(mn^2 + n^3)$, with $n$ being the number of filters, and $m$ the number of samples. For example, for ImageNet, $m$=1.2M, and assuming a VGG-16 architecture with layers of size $n=$ 64, 128, 256, 512, and 4096, the time to compute PFA per layer is roughly 1.24s, 2.8s, 4.6s, 9.3s, and 127.5s respectively (single CPU @ 2.30GHz). The complexity of the filter selection depends only on the layer size. In the worst case the complexity is $O(rn^2)$, where $r$ is the number of filters to remove. 

Considering that PFA has to run only once at the end of the training step, the time consumed by PFA is negligible compared to the whole training time. In exchange for this negligible additional time, PFA provides the long-term benefit of a smaller footprint and faster inference, which, in the lifetime of a deployed network, including re-training when new data becomes available, will quickly surpass the time initially required by PFA.

Once the eigenvalue set $\bL$ is computed for all layers, generating PFA recipes with different energy values is extremely fast. Hence, the threshold in the PFA-En strategy can conveniently be replaced with the target model size or FLOPs. PFA-En can then be computed iteratively with decreasing energy thresholds until the requested size is achieved. This seemingly small change in the interaction with the user is a great benefit since it is often difficult to relate algorithms parameters to practical characteristics (such as the size) of the final model.

\subsection{Discussion}
\label{sec:limitations}

All state-of-the-art techniques analyzed achieve great results in term of maintaining accuracy and reducing memory footprint and FLOPs. In general, even without training hyper-parameter search, PFA yields competitive results. We believe, however, that the biggest advantage of PFA is its simplicity and efficacy compared to other techniques.

Often state-of-the-art algorithms are not adopted because of the high friction required for their application. For example, \VI requires the user to modify the network to perform a sampling step during the forward pass, \FG requires an optimization problem to be solved for each layer  to decompose a convolutional layer into a group of smaller operations that approximate the output of the full layer, and \NS requires the training protocol to be modified to induce sparsity in the full model. PFA is based on an intuitive idea: remove filters that produce correlated responses. This makes its implementation, application and adoption straightforward.

PFA does not need to modify any loss function, unlike~\NS and \VI, a potential second barrier to adoption. This makes it attractive because known hyper-parameters can be used for the full model, and also makes PFA deployable as a service: given a full model and a dataset, PFA can provide an initialized smaller model (without needing to know the loss function). Furthermore, while intuitively one might expect that techniques that modify the loss function should obtain better results (since the compression aspect is included in the optimization) our experiments did not show a consistent benefit compared to PFA.

All state-of-the-art techniques analyzed require user defined parameters that need additional tuning and are difficult to relate the FLOPs or the size of the compressed model. \FP needs a threshold to decide if the $\ell_1$-norm of a filter is small enough to be pruned. This process is non-trivial and requires the user to choose the compression thresholds based on a pre-analysis that provides insight on the sensitivity of each layer to pruning. \NS~has two crucial parameters: the weight of the sparsity regularizer, and the percentage of filters to be pruned. The weight has a direct impact on the induced sparsity, however, there is no intuitive way to set this parameter and it needs to be tuned for each architecture and dataset. In addition, setting the same percentage of filters to be pruned at each layer for the whole network ignores the relative effect of those filters on the accuracy and the footprint of the network. \VI requires a parameter to control the influence of the information bottleneck term, which is related to the compression achieved. \FG requires a parameter that defines the compression ratio for each layer. In both algorithms, the tuning is different depending both on the network-dataset combination and the layer depth. From the results there seems to be no intuitive way to set this parameter other than by trial and error. In contrast, PFA requires a single intuitive parameter (for example the desired model size in PFA-En), or it is parameter-free (PFA-KL).

Lastly, compared to techniques based on weight analysis, such as~\FP, PFA is based on the responses of a layer. This means that different datasets used for the PFA analysis leads to different and specialized models, as we will describe in~\Sec~\ref{sec:domain}, which makes PFA a suitable candidate for the task of simultaneous compression and domain adaptation.

\subsection{Simultaneous compression and domain adaptation using PFA}
\label{sec:domain}

\begin{figure*}[tb]
		\centering
		\subfigure[Domain adaptation from CIFAR-100]{\label{fig:domain_acc_c100}\includegraphics[width=0.49\textwidth]{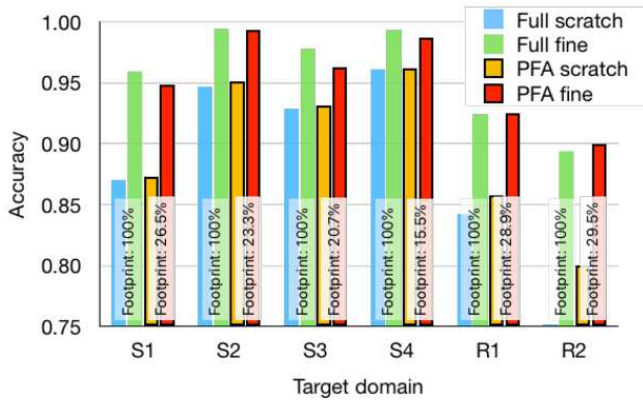}}
		\subfigure[PFA recipes starting from CIFAR-100]{\label{fig:domain_recipes_c100}\includegraphics[width=0.49\textwidth]{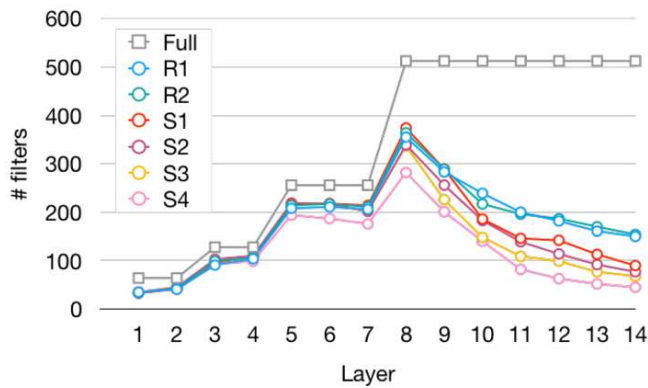}}
	\caption	{Domain adaption from CIFAR-100. (a) {\emph{PFA fine}} matches the accuracy of {\emph{Full fine}} while using architectures more than 4x smaller. {\emph {PFA fine}} significantly outperforms the full model trained from scratch {\emph{Full scratch}}. The vertical percentage labels show the PFA compression ratio. In (b) recipes for VGG-16 trained on CIFAR-100 using PFA-KL with data from different target domains. Note how PFA exploits the knowledge of the target domain, creating different recipes adapted to the task complexity.}\label{fig:domain}
\end{figure*}

By compressing networks based on their responses, rather than their weights, the compressed networks become specialized for the target domain at hand: depending on the dataset used to generate the responses the compressed architecture will change. In this section, we show some examples of how PFA modifies the same original architecture differently to adapt to different target datasets, while taking advantage of the original training.

Let us denote the initial domain used for training as $\mathcal{D}_A$; in this test $\mathcal{D}_A$ is CIFAR-100. We denote the domain used for PFA as $\mathcal{D}_Z$. We generate different $\mathcal{D}_Z$s by  randomly sampling classes out of the original 100 classes contained in CIFAR-100. We generate two targets $\mathcal{D}_Z$ of 10 classes each (R1 and R2), and four targets $\mathcal{D}_Z$  of 2 classes each (S1, S2, S3, and S4). Refer to App.~\ref{app:domain} for a detailed explanation of the target domains used.

For each adaptation $\mathcal{D}_A \rightarrow \mathcal{D}_Z$ we run the following experiments using a VGG-16 model:
\begin{itemize}
\item \textbf{Full scratch}: Train from scratch on domain $\mathcal{D}_Z$.
\item \textbf{Full fine}: Train from scratch on domain $\mathcal{D}_A$ and fine-tune on $\mathcal{D}_Z$.
\item \textbf{PFA scratch}: Train from scratch on domain $\mathcal{D}_A$, run PFA-KL on domain $\mathcal{D}_Z$ and train the compressed architecture from scratch on $\mathcal{D}_Z$.
\item \textbf{PFA fine}: Train from scratch on domain $\mathcal{D}_A$, run PFA-KL on domain $\mathcal{D}_Z$ and train the compressed architecture using filter selection on $\mathcal{D}_Z$.
\end{itemize}

The results in \Fig~\ref{fig:domain_acc_c100} show how the \emph{PFA fine} strategy (red bars) performs similarly to the full fined tuned model (\emph{Full fine}, green bars), while obtaining models more than 4 times smaller. %
Moreover, the \emph{PFA fine} strategy significantly outperforms the full model trained from scratch on the target domain (\emph{Full scratch}, blue bars).

The compressed architectures generated by PFA, \Fig~\ref{fig:domain_recipes_c100}, are different depending on the complexity of the final task. Note how PFA obtains architectures with more filters for the 10 class subsets (R1 and R2) than for the 2 class subset (S1, S2, S3, and S4). Even among the 2 class subset, there is a small variation in the final architecture, reflecting the different level of difficulty to distinguish between the two target classes. 

These results show how by analyzing the responses rather than the weights, PFA is able to compress a network while specializing it to different domains. %

\section{Conclusions}
Two effective, and yet easy to implement techniques for the compression of neural networks using Principal Filter Analysis are presented: PFA-En and PFA-KL. These techniques exploit the correlation of filter responses within layers to compress networks without compromising accuracy. PFA can be applied to the output response of any layer with no knowledge of the training procedure or the loss function.  %
Our tests show that PFA is competitive with state-of-the-art approaches while removing adoption barriers thanks to its practical implementation, intuitive philosophy and ease of use. PFA-KL is parameter free, and PFA-En has only a single intuitive parameter: the energy to be preserved in each layer or a desired network characteristic (such as a target model size or FLOPs).

The flexibility of PFA makes it applicable to a wide variety of architectures that we would like to investigate in future: recurrent neural networks, models with attention, and even word embedding.

{
\bibliographystyle{ieee}
\bibliography{bibliography/deep_learning}

\begin{thebibliography}{10}\itemsep=-1pt

\bibitem{Aghasi:NIPS:2017}
A.~Aghasi, A.~Abdi, N.~Nguyen, and J.~Romberg.
\newblock Net-trim: Convex pruning of deep neural networks with performance
  guarantee.
\newblock In {\em NIPS}, pages 3180--3189, 2017.

\bibitem{Alvarez:NIPS:2017}
J.~M. Alvarez and M.~Salzmann.
\newblock Compression-aware training of deep networks.
\newblock volume abs/1711.02638, 2017.

\bibitem{Arora:ICML:2018}
S.~Arora, N.~Cohen, and E.~Hazan.
\newblock On the optimization of deep networks: Implicit acceleration by
  overparameterization.
\newblock In {\em NIPS}, 2018.

\bibitem{Ba:NIPS:2014}
L.~J. Ba and R.~Caurana.
\newblock Do deep nets really need to be deep?
\newblock In {\em NIPS}, 2014.

\bibitem{Bucilua:SIGKDD:2006}
C.~Bucilua, R.~Caruana, and A.~Niculescu-Mizil.
\newblock Model compression.
\newblock In {\em SIGKDD}, pages 535--541, 2006.

\bibitem{Carreira:CVPR:2018}
M.~A. Carreira-Perpinan and Y.~Idelbayev.
\newblock “learning-compression” algorithms for neural net pruning.
\newblock In {\em CVPR}, 2018.

\bibitem{Chen:ICLR:2016}
T.~Chen, I.~Goodfellow, and J.~Shlens.
\newblock Net2net: Accelerating learning via knowledge transfer.
\newblock In {\em ICLR}, 11 2016.

\bibitem{Tai:ICLR:2016}
T.~Cheng, X.~Tong, W.~Xiaogang, and W.~E.
\newblock Convolutional neural networks with low-rank regularization.
\newblock In {\em ICLR}, 2016.

\bibitem{Cogswell:ICLR:2016}
M.~Cogswell, F.~Ahmed, R.~B. Girshick, L.~Zitnick, and D.~Batra.
\newblock Reducing overfitting in deep networks by decorrelating
  representations.
\newblock In {\em ICLR}, 2016.

\bibitem{Cohen:CORR:2018}
G.~Cohen, G.~Sapiro, and R.~Giryes.
\newblock {DNN} or k-nn: That is the generalize vs. memorize question.
\newblock {\em CoRR}, 2018.

\bibitem{Dai:ICML:2018}
B.~Dai, C.~Zhu, B.~Guo, and D.~Wipf.
\newblock Compressing neural networks using the variational information
  bottleneck.
\newblock In {\em ICML}, 2018.

\bibitem{ImageNet}
J.~Deng, W.~Dong, R.~Socher, L.-J. Li, K.~Li, and L.~Fei-Fei.
\newblock Imagenet: A large-scale hierarchical image database.
\newblock In {\em CVPR}, 2009.

\bibitem{Denil:NIPS:2013}
M.~Denil, B.~Shakibi, L.~Dinh, M.~Ranzato, and N.~de~Freitas.
\newblock Predicting parameters in deep learning.
\newblock In {\em NIPS}, pages 2148--2156, 2013.

\bibitem{Denton:NIPS:2014}
E.~L. Denton, W.~Zaremba, J.~Bruna, Y.~LeCun, and R.~Fergus.
\newblock Exploiting linear structure within convolutional networks for
  efficient evaluation.
\newblock In {\em NIPS}, pages 1269--1277, 2014.

\bibitem{tensorflow2015-whitepaper}
M.~A. et~al.
\newblock {TensorFlow}: Large-scale machine learning on heterogeneous systems,
  2015.
\newblock Software available from tensorflow.org.

\bibitem{Han:ICLR:2016}
S.~Han, H.~Mao, and W.~J. Dally.
\newblock Deep compression: Compressing deep neural networks with pruning,
  trained quantization and huffman coding.
\newblock In {\em ICLR}, 2016.

\bibitem{Han:NIPS:2015}
S.~Han, J.~Pool, J.~Tran, and W.~J. Dally.
\newblock Learning both weights and connections for efficient neural networks.
\newblock In {\em NIPS}, 2016.

\bibitem{Hassibi:NIPS:1993}
B.~Hassibi, D.~G. Stork, and G.~Wolff.
\newblock Optimal brain surgeon: Extensions and performance comparisons.
\newblock In {\em NIPS}, pages 263--270, 1993.

\bibitem{He:CVPR:2016}
K.~He, X.~Zhang, S.~Ren, and J.~Sun.
\newblock Deep residual learning for image recognition.
\newblock In {\em CVPR}, pages 770--778, 2016.

\bibitem{He:ECCV:2018}
X.~He and J.~Cheng.
\newblock Learning compression from limited unlabeled data.
\newblock In {\em ECCV}, 2018.

\bibitem{He:ICCV:2017}
Y.~He, X.~Zhang, and J.~Sun.
\newblock Channel pruning for accelerating very deep neural networks.
\newblock In {\em ICCV}, 2017.

\bibitem{Hinton:NIPS:2014}
G.~Hinton, J.~Dean, and O.~Vinyals.
\newblock Distilling the knowledge in a neural network.
\newblock In {\em NIPS}, pages 1--9, 03 2014.

\bibitem{PCA:1933}
H.~Hotelling.
\newblock Analysis of a complex of statistical variables into principal
  components.
\newblock {\em Journal of Educational Psychology}, 24:417--441 and 498--520,
  1933.

\bibitem{Jaderberg:BMVC:2014}
M.~Jaderberg, A.~Vedaldi, and A.~Zisserman.
\newblock Speeding up convolutional neural networks with low rank expansions.
\newblock In {\em BMVC}, 2014.

\bibitem{CIFAR}
A.~Krizhevsky.
\newblock Learning multiple layers of features from tiny images.
\newblock Master's thesis, 2009.

\bibitem{kullback1951}
S.~Kullback and R.~Leibler.
\newblock On information and sufficiency.
\newblock {\em Ann. Math. Statist.}, 22(1):79--86, 03 1951.

\bibitem{Lebedev:ICLR:2015}
V.~Lebedev, Y.~Ganin, M.~Rakhuba, I.~V. Oseledets, and V.~S. Lempitsky.
\newblock Speeding-up convolutional neural networks using fine-tuned
  cp-decomposition.
\newblock In {\em ICLR}, 2014.

\bibitem{Lecun:NIPS:1990}
Y.~LeCun, J.~S. Denker, and S.~A. Solla.
\newblock Optimal brain damage.
\newblock In {\em NIPS}, pages 598--605, 1990.

\bibitem{Li:ICLR:2017}
H.~Li, A.~Kadav, I.~Durdanovic, H.~Samet, and H.~P. Graf.
\newblock Pruning filters for efficient convnets.
\newblock In {\em ICLR}, 2017.

\bibitem{Liu:ICCV:2017}
Z.~Liu, J.~Li, Z.~Shen, G.~Huang, S.~Yan, and C.~Zhang.
\newblock Learning efficient convolutional networks through network slimming.
\newblock In {\em ICCV}, 2017.

\bibitem{Livni:NIPS:2014}
R.~Livni, S.~Shalev-Shwartz, and O.~Shamir.
\newblock On the computational efficiency of training neural networks.
\newblock In {\em NIPS}, 2014.

\bibitem{Long:CVPR:2015}
J.~Long, E.~Shelhamer, and T.~Darrell.
\newblock Fully convolutional networks for semantic segmentation.
\newblock In {\em CVPR}, 2015.

\bibitem{Luo:ICCV:2017}
J.-H. Luo, J.~Wu, and W.~Lin.
\newblock Thinet: A filter level pruning method for deep neural network
  compression.
\newblock In {\em ICCV}, 2017.

\bibitem{Ma:Corr:2018}
M.~Ma, H.~Pouransari, D.~Chao, S.~Adya, S.~A. Serrano, Y.~Qin, D.~Gimnicher,
  and D.~Walsh.
\newblock Democratizing production-scale distributed deep learning.
\newblock {\em CoRR}, abs/1811.00143, 2018.

\bibitem{Masana:ICCV:2017}
M.~Masana, J.~van~de Weijer, L.~Herranz, A.~D. Bagdanov, and J.~M.
  {\'{A}}lvarez.
\newblock Domain-adaptive deep network compression.
\newblock In {\em ICCV}, 2017.

\bibitem{Molchanov:ICLR:17}
P.~Molchanov, S.~Tyree, T.~Karras, T.~Aila, and J.~Kautz.
\newblock Pruning convolutional neural networks for resource efficient
  inference.
\newblock In {\em ICLR}, 2017.

\bibitem{Nakkiran:IS:2015}
P.~Nakkiran, R.~Alvarez, R.~Prabhavalkar, and C.~Parada.
\newblock Compressing deep neural networks using a rank-constrained topology.
\newblock pages 1473--1477, 2015.

\bibitem{Pearson:RSL:1895}
K.~Pearson.
\newblock Notes on regression and inheritance in the case of two parents.
\newblock In {\em Royal Society of London}, volume~58, pages 240--242, 1895.

\bibitem{Peng:ECCV:2018}
B.~Peng, W.~Tan, Z.~Li, S.~Zhang, D.~Xie, and S.~Pu.
\newblock Extreme network compression via filter group approximation.
\newblock In {\em ECCV}, 2018.

\bibitem{Prabhavalkar:ICASSP:2016}
R.~Prabhavalkar, O.~Alsharif, A.~Bruguier, and I.~McGraw.
\newblock On the compression of recurrent neural networks with an application
  to {LVCSR} acoustic modeling for embedded speech recognition.
\newblock In {\em ICASSP}, pages 5970--5974, 2016.

\bibitem{Nguyen:ICML:2017}
N.~Quynh and H.~Matthias.
\newblock The loss surface of deep and wide neural networks.
\newblock In {\em ICML}, 2017.

\bibitem{Rastegari:ECCV:2016}
M.~Rastegari, V.~Ordonez, J.~Redmon, and A.~Farhadi.
\newblock Xnor-net: Imagenet classification using binary convolutional neural
  networks.
\newblock In {\em ECCV}, 2016.

\bibitem{Rodriguez:ICLR:2017}
P.~Rodr{\'{\i}}guez, J.~Gonz{\`{a}}lez, G.~Cucurull, J.~M. Gonfaus, and F.~X.
  Roca.
\newblock Regularizing cnns with locally constrained decorrelations.
\newblock In {\em ICLR}, 2017.

\bibitem{Romero:ICRL:2015}
A.~Romero, N.~Ballas, S.~E. Kahou, A.~Chassang, C.~Gatta, and Y.~Bengio.
\newblock Fitnets: Hints for thin deep nets.
\newblock In {\em ICLR}, 2015.

\bibitem{Rubner:ICCV:1998}
Y.~Rubner, C.~Tomasi, and L.~J. Guibas.
\newblock A metric for distributions with applications to image databases.
\newblock In {\em ICCV}, 1998.

\bibitem{Simonyan:ICLR:2015}
K.~Simonyan and A.~Zisserman.
\newblock Very deep convolutional networks for large-scale image recognition.
\newblock In {\em ICLR}, 2015.

\bibitem{Srinivas:BMVC:2015}
S.~Srinivas and R.~V. Babu.
\newblock Data-free parameter pruning for deep neural networks.
\newblock In {\em BMVC}, pages 31.1--31.12, 2015.

\bibitem{Tung:CVPR:2018}
F.~Tung and G.~Mori.
\newblock Clip-q: Deep network compression learning by in-parallel
  pruning-quantization.
\newblock In {\em CVPR}, 2018.

\bibitem{Wen:NIPS:2016}
W.~Wen, C.~Wu, Y.~Wang, Y.~Chen, and H.~Li.
\newblock Learning structured sparsity in deep neural networks.
\newblock In {\em NIPS}, pages 2074--2082, 2016.

\bibitem{Wen:ICCV:2017}
W.~Wen, C.~Xu, C.~Wu, Y.~Wang, Y.~Chen, and H.~Li.
\newblock Coordinating filters for faster deep neural networks.
\newblock In {\em ICCV}, 2017.

\bibitem{Jiaxiang:CVPR:2016}
J.~Wu, C.~Leng, Y.~Wang, Q.~Hu, and J.~Cheng.
\newblock Quantized convolutional neural networks for mobile devices.
\newblock In {\em CVPR}, 2016.

\bibitem{Wu:ICML:2018}
J.~Wu, Y.~Wang, Z.~Wu, Z.~Wang, A.~Veeraraghavan, and Y.~Lin.
\newblock Deep k-means: Re-training and parameter sharing with harder cluster
  assignments for compressing deep convolutions.
\newblock In {\em ICML}, 2018.

\bibitem{Xue:ICASSP:2013}
J.~Xue, J.~Li, and Y.~Gong.
\newblock Restructuring of deep neural network acoustic models with singular
  value decomposition.
\newblock In {\em ICASSP}, pages 6359--6363, 2013.

\bibitem{Yu:CVPR:2018}
R.~Yu, A.~Li, C.-F. Chen, J.-H. Lai, V.~I. Morariu, X.~Han, M.~Gao, C.-Y. Lin,
  and L.~S. Davis.
\newblock Nisp: Pruning networks using neuron importance score propagation.
\newblock In {\em CVPR}, 2018.

\bibitem{Yu:CVPR:2017}
X.~Yu, T.~Liu, X.~Wang, and D.~Tao.
\newblock On compressing deep models by low rank and sparse decomposition.
\newblock In {\em CVPR}, pages 7370--7379, 2017.

\bibitem{Zagoruyko:TORCH:2015}
S.~Zagoruyko.
\newblock 92.45\% on cifar-10 in torch.
\newblock \url{https://github.com/szagoruyko/cifar.torch}, 2015.

\bibitem{Zhang:ECCV:2018}
T.~Zhang, S.~Ye, K.~Zhang, J.~Tang, W.~Wen, M.~Fardad, and Y.~Wang.
\newblock A systematic dnn weight pruning framework using alternating direction
  method of multipliers.
\newblock In {\em ECCV}, 2018.

\end{thebibliography}
}

\newpage
\newpage
\onecolumn

\appendix
\appendixpage

\section{Details on the domain adaptation experiments}\label{app:domain}
In this section the exact classes used for the domain adaptation experiments shown in \Sec~\ref{sec:domain} are presented. In \Sec~\ref{sec:domain} the initial domain $\mathcal{D}_A$ is equal to the whole CIFAR-100 dataset. We then use different subsets of CIFAR-100 to generate target domains $\mathcal{D}_Z$. Here we are specifying exactly which classes are used for the different 
$\mathcal{D}_Z$s:

\begin{itemize}
\item \textbf{R1}: aquarium\_fish, butterfly, cloud, elephant, mountain, palm\_tree, poppy, snail, squirrel, wardrobe;
\item \textbf{R2}: baby, camel, cockroach, flatfish, mouse, pear, porcupine, snake, sunflower, whale;
\item \textbf{S1}: bowl, snail;
\item \textbf{S2}: aquarium\_fish, boy;
\item \textbf{S3}: bee, raccoon;
\item \textbf{S4}: train, tulip.
\end{itemize}
\newpage

\pagebreak
\section{Compression Results}
\label{app:details}

In this section for each pair architecture-dataset that we used in the experiments presented in \Sec~\ref{sec:quantitative} we report the accuracy achieved by the full model used for pruning, and for each PFA recipe we provide the change in the accuracy and the percentage of the trainable variables and FLOPs of the pruned model with respect to the full model. The number of trainable variables is computed by summing the products of the shape of all trainable variables returned by TensorFlow~\cite{tensorflow2015-whitepaper} 1.4.0 {\tt trainable.variable()} API. The FLOPs are computed by running the TensorFlow 1.4.0 profiler. Furthermore, when available, we report the same details for state-of-the-art work that we used for comparison. Results for VGG-16 with CIFAR-10 and CIFAR-100 are in Tabs.~\ref{tab:vgg_c10} and~\ref{tab:vgg_c100}. Results for ResNet-56 with CIFAR-10 and CIFAR-100 are in Tabs.~\ref{tab:res56_c10} and~\ref{tab:res56_c100}. Results for VGG-16 with ImageNet (top-1 and top-5) are in Tabs.~\ref{tab:vgg_imagenet_top1} and~\ref{tab:vgg_imagenet_top5}. Results for ResNet-18 with ImageNet (top-1 and top-5) are in Tabs.~\ref{tab:resnet_imagenet_top1} and~\ref{tab:resnet_imagenet_top5}.

\begin{figure}[H]
  \centering
  \subfigure[VGG-16 on ImageNet]{\label{fig:vgg16_imagenet} \includegraphics[width=0.49\textwidth]{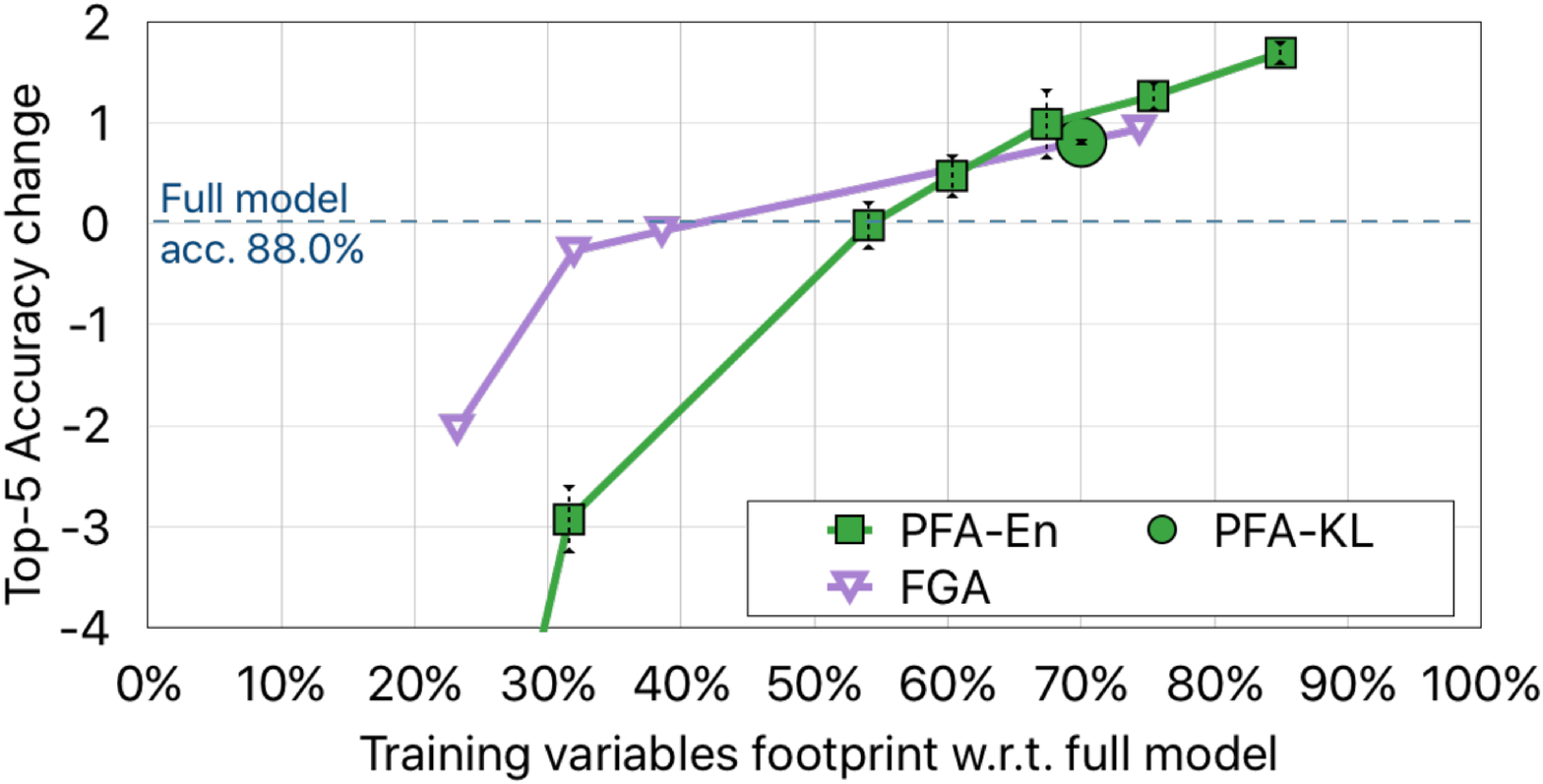}}
  \subfigure[ResNet-34 on ImageNet]{\label{fig:resnet34_imagenet} \includegraphics[width=0.49\textwidth]{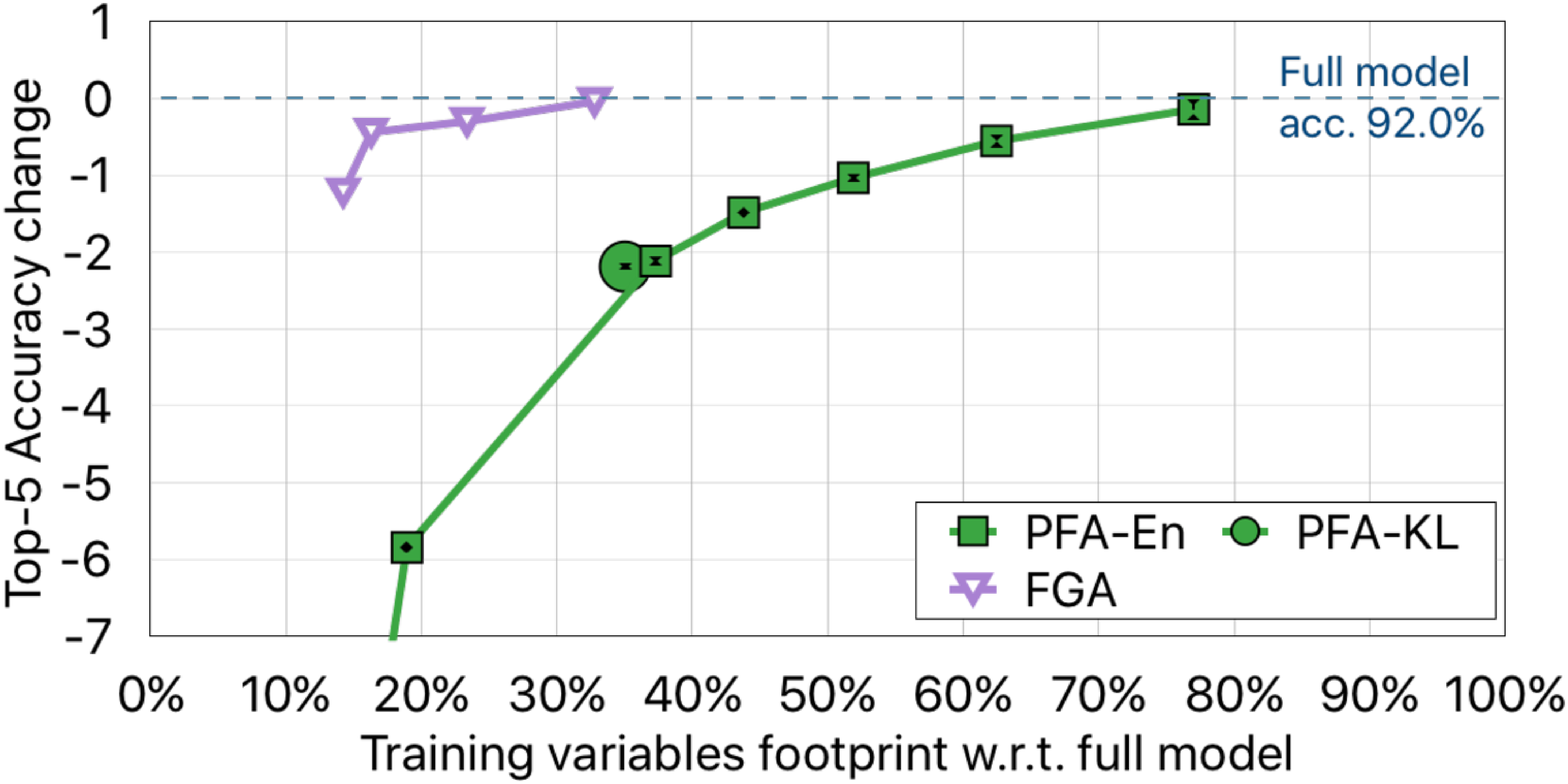}}
  \caption{Top-5 results on ImageNet. 
    Accuracy change in the $y$ axis is reported in percentage points.\label{fig:top-5}}
\end{figure}

In \Sec~\ref{sec:quantitative} we presented, among others, the Top-1 accuracy of PFA applied to VGG-16 and ResNet-34 on ImageNet (\Fig~\ref{fig:imagenet_top1}). Here, in \Fig~\ref{fig:top-5} we also provide the Top-5 results for the same experiment.

\begin{table}[H]
  \centering
  \caption{VGG-16 with CIFAR-10.}
  \begin{tabular}{l|ccc}
    VGG-16 &  Accuracy & \% Trainable Var. & \% FLOPs \\
  \toprule
  \toprule
  Full Model & 92.07\% & 100\% & 100\% \\
             & $\Delta$ Accuracy & \% Trainable Var. & \% FLOPs \\
  \toprule
  \toprule
  PFA-En 0.99 & 0.31 & 18.25\% & 51.98\%\\
  PFA-En 0.98 & 0.35 & 12.70\% & 42.27\%\\
  PFA-En 0.97 & 0.14 & 10.03\% & 36.29\%\\
  PFA-En 0.96 & -0.01 & 8.33\% & 31.67\%\\
  PFA-En 0.95 & -0.05 & 7.12\% & 28.21\%\\
  PFA-En 0.93 & -0.27 & 5.45\% & 22.56\%\\
  PFA-En 0.90 & -0.97 & 3.87\% & 16.61\%\\
  PFA-En 0.80 & -3.97 & 1.38\% & 6.10\%\\
  \midrule
  PFA-KL & 0.244 & 19.63\% & 41.70\% \\
  \midrule
  \FP~\cite{Li:ICLR:2017} & 0.15 & 36.00\% & 65.81\%\\  
  \NS~\cite{Liu:ICCV:2017} & 0.14 & 11.48\% & 49.06\%\\  
  \VI~\cite{Dai:ICML:2018} & 0.81 & 5.79\% & N.A.\\  
  \bottomrule  
  \end{tabular}
  \label{tab:vgg_c10}
\end{table}
 
\begin{table}[H]
  \centering
  \caption{VGG-16 with CIFAR-100.}
  \begin{tabular}{l|ccc}
  VGG-16 & Accuracy & \% Trainable Var. & \% FLOPs \\
    \toprule
     \toprule
  Full Model & 68.36\% & 100\% & 100\% \\
          & $\Delta$ Accuracy & \% Trainable Var. & \% FLOPs \\
           \toprule
     \toprule
  PFA-En 0.99 & 1.68 & 43.15\% & 70.97\%\\
  PFA-En 0.98 & 1.41 & 33.14\% & 59.97\%\\
  PFA-En 0.97 & 1.21 & 27.13\% & 52.13\%\\
  PFA-En 0.96 & 0.78 & 22.72\% & 45.91\% \\
  PFA-En 0.95 & 0.50 & 19.29\% & 40.31\%\\
  PFA-En 0.93 & -0.27 & 14.39\% & 32.06\%\\
  PFA-En 0.90 & -2.37 & 9.66\% & 23.16\%\\
  PFA-En 0.80 & -8.71 & 3.01\% & 8.12\%\\
  \midrule
  PFA-KL & 1.40 & 41.91\% & 55.29\% \\
  \midrule
  \NS~\cite{Liu:ICCV:2017} & 0.22 & 24.90\% & 62.86\%\\  
  \VI~\cite{Dai:ICML:2018} & 1.17 & 15.08\% & N.A.\\  
  \FG/A~\cite{Peng:ECCV:2018} & 0.39 & 39.67\% & 59.97\% \\
  \FG/B~\cite{Peng:ECCV:2018} & -0.02 & 18.54\% & 26.60\%\\
  \FG/C~\cite{Peng:ECCV:2018} & -0.57 & 13.20\% & 15.18\%\\
  \FG/D~\cite{Peng:ECCV:2018} & -1.93 & 11.31\% & 11.42\%\\
  \bottomrule  
  \bottomrule  
  \end{tabular}
  \label{tab:vgg_c100}
\end{table}

\begin{table}[H]
  \centering
  \caption{ResNet-56 with CIFAR-10.}
  \begin{tabular}{l|ccc}
    ResNet-56 & Accuracy & \% Trainable Var. & \% FLOPs \\
    \toprule
    \toprule
  Full Model & 93.15\% & 100\% & 100\% \\
 & $\Delta$ Accuracy & \% Trainable Var. & \% FLOPs \\
      \toprule
    \toprule
  PFA-En 0.99 & -0.20 & 72.60\% & 80.69\% \\
  PFA-En 0.98 & -0.18 & 61.89\% & 70.04\% \\
  PFA-En 0.97 & -0.80 & 54.10\% & 61.85\% \\
  PFA-En 0.96 & -0.70 & 48.27\% & 55.46\% \\
  PFA-En 0.95 & -1.01 & 43.36\% & 49.51\% \\
  PFA-En 0.93 & -1.25 & 36.11\% & 41.61\% \\
  PFA-En 0.90 & -1.81 & 27.42\% & 28.08\% \\
  PFA-En 0.80 & -4.10 & 12.57\% & 14.56\%\\
  \midrule
  PFA-KL & -0.65 & 59.97\% & 61.57\% \\
  \midrule
  \FP/A~\cite{Li:ICLR:2017} & 0.06 & 90.59\% & 89.60\%\\  
  \FP/B~\cite{Li:ICLR:2017} & 0.02 & 85.88\% & 72.72\%\\  
  \bottomrule  
   \bottomrule  
  \end{tabular}
  \label{tab:res56_c10}
\end{table}
 
\begin{table}[H]
  \centering
  \caption{ResNet-56 with CIFAR-100.}
  \begin{tabular}{l|ccc}
  ResNet-56 & Accuracy & \% Trainable Var. & \% FLOPs \\
    \toprule
       \toprule
  Full Model & 70.92 & 100\% & 100\% \\
  & $\Delta$ Accuracy & \% Trainable Var. & \% FLOPs \\
    \toprule
       \toprule
  PFA-En 0.99 & -1.86 & 90.28\% & 89.46\%\\
  PFA-En 0.98 & -1.68 & 81.95\% & 79.36\%\\
  PFA-En 0.97 & -2.33 & 74.92\% & 71.54\%\\
  PFA-En 0.96 & -2.75 & 68.50\% & 64.80\%\\
  PFA-En 0.95 & -3.50 & 62.76\% & 59.20\%\\
  PFA-En 0.93 & -4.29 & 35.87\% & 41.61\%\\
  PFA-En 0.90 & -5.16 & 27.24\% & 29.18\%\\
  PFA-En 0.80 & -9.70 & 19.58\% & 16.76\%\\
  \midrule
  PFA-KL & -2.97 & 74.00\% & 65.60\%\\
  \midrule
  -- & & & \\  
  \bottomrule  
   \bottomrule  
  \end{tabular}
  \label{tab:res56_c100}
\end{table}

\begin{table}[H]
  \centering
  \caption{VGG-16-Conv Top-1  with ImageNet.}
  \begin{tabular}{l|ccc}
  VGG-16 Top-1 & Accuracy & \% Model Size & \% FLOPs \\
    \toprule
    \toprule
  Full Model & 67.73\% & 100\% & 100\% \\
    & $\Delta$ Accuracy & \% Model Size & \% FLOPs \\
    \toprule
    \toprule
  PFA-En 0.99 & 2.80 & 84.98\% & 77.31\%\\
  PFA-En 0.98 & 1.97 & 75.40\% & 66.90\%\\
  PFA-En 0.97 & 1.59 & 67.43\% & 58.75\%\\
  PFA-En 0.96 & 0.90 & 60.32\% & 52.03\%\\
  PFA-En 0.95 & 0.02 & 54.06\% & 46.14\%\\
  PFA-En 0.90 & -5.02 &  31.58\%& 26.46\%\\
  PFA-En 0.85 & -20.94& 10.32\% & 8.72\%\\
  \midrule
  PFA-KL & 1.06 & 70.03\% & 53.57\% \\
  \midrule
  \FG/A~\cite{Peng:ECCV:2018} & 0.82 & 74.37\% & 43.01\% \\
  \FG/B~\cite{Peng:ECCV:2018} & -0.28 & 38.55\% & 22.14\%\\
  \FG/C~\cite{Peng:ECCV:2018} & -1.06 & 31.95\% & 19.67\%\\
  \FG/D~\cite{Peng:ECCV:2018} & -3.49 & 23.18\% & 15.16\%\\
  \bottomrule  
  \end{tabular}
  \label{tab:vgg_imagenet_top1}
\end{table}

\begin{table}[H]
  \centering
  \caption{VGG-16-Conv Top-5  with ImageNet.}
  \begin{tabular}{l|ccc}
  VGG-16 Top-5 & Accuracy & \% Model Size & \% FLOPs \\
    \toprule
    \toprule
  Full Model & 88.03\% & 100\% & 100\% \\
    & $\Delta$ Accuracy & \% Model Size & \% FLOPs \\
    \toprule
    \toprule
  PFA-En 0.99 & 1.70 & 84.98\% & 77.31\%\\
  PFA-En 0.98 & 1.27 & 75.40\% & 66.90\%\\
  PFA-En 0.97 & 0.99 & 67.43\% & 58.75\%\\
  PFA-En 0.96 & 0.47 & 60.32\% & 52.03\%\\
  PFA-En 0.95 & -0.02 & 54.06\% & 46.14\%\\
  PFA-En 0.90 & -2.93 &  31.58\%& 26.46\%\\
  PFA-En 0.85 & -15.19& 10.32\% & 8.72\%\\
  \midrule
  PFA-KL      & 0.81 & 70.03\% & 53.57\% \\
  \midrule
  \FG/A~\cite{Peng:ECCV:2018} & 0.94 & 74.37\% & 43.01\% \\
  \FG/B~\cite{Peng:ECCV:2018} & -0.07 & 38.55\% & 22.14\%\\
  \FG/C~\cite{Peng:ECCV:2018} & -0.27 & 31.95\% & 19.67\%\\
  \FG/D~\cite{Peng:ECCV:2018} & -2.03 & 23.18\% & 15.16\%\\
  \bottomrule  
  \end{tabular}
  \label{tab:vgg_imagenet_top5}
\end{table}

\begin{table}[H]
  \centering
  \caption{ResNet-34 Top-1 with ImageNet.}
  \begin{tabular}{l|ccc}
  ResNet-34 Top-1 & Accuracy & \% Model Size & \% FLOPs \\
    \toprule
    \toprule
  Full Model & 74.49\% & 100\% & 100\% \\
    & $\Delta$ Accuracy & \% Model Size & \% FLOPs \\
    \toprule
    \toprule
  PFA-En 0.99 & -0.27 & 77.02\% & 76.00\%\\
  PFA-En 0.98 & -1.08 & 62.46\% & 65.39\%\\
  PFA-En 0.97 & -1.85 & 51.88\% & 57.14\%\\
  PFA-En 0.96 & -2.71 & 43.79\% & 50.47\%\\
  PFA-En 0.95 & -3.83 & 37.30\% & 44.73\%\\
  PFA-En 0.90 & -9.78 & 18.94\% & 26.02\%\\
  PFA-En 0.85 & -29.05& 6.37\% & 9.45\%\\
  \midrule
  PFA-KL & -4.04 & 35.02\% & 48.21\% \\
  \midrule
  \FP/A~\cite{Li:ICLR:2017} & -0.67 & 92.13\% & 84.62\%\\  
  \FP/B~\cite{Li:ICLR:2017} & -1.06 & 89.35\% & 75.82\%\\  
  \FP/C~\cite{Li:ICLR:2017} & -0.75 & 93.06\% & 92.58\%\\  
  \FG/A~\cite{Peng:ECCV:2018} & -0.35 & 32.78\% & 54.37\% \\
  \FG/B~\cite{Peng:ECCV:2018} & -1.02 & 23.37\% & 35.25\%\\
  \FG/C~\cite{Peng:ECCV:2018} & -1.70 & 16.30\% & 19.67\%\\
  \FG/D~\cite{Peng:ECCV:2018} & -3.03 & 14.23\% & 15.16\%\\
  \bottomrule  
  \end{tabular}
  \label{tab:resnet_imagenet_top1}
\end{table} 
 
\begin{table}[H]
  \centering
  \caption{ResNet-34 Top-5 with ImageNet.}
  \begin{tabular}{l|ccc}
  ResNet-34 Top-5 & Accuracy & \% Model Size & \% FLOPs \\
    \toprule
    \toprule
  Full Model & 91.99\% & 100\% & 100\% \\
    & $\Delta$ Accuracy & \% Model Size & \% FLOPs \\
    \toprule
    \toprule
  PFA-En 0.99 & -0.14 & 77.02\% & 76.00\%\\
  PFA-En 0.98 & -0.55 & 62.46\% & 65.39\%\\
  PFA-En 0.97 & -1.03 & 51.88\% & 57.14\%\\
  PFA-En 0.96 & -1.49 & 43.79\% & 50.47\%\\
  PFA-En 0.95 & -2.12 & 37.30\% & 44.73\%\\
  PFA-En 0.90 & -5.84 & 18.94\% & 26.02\%\\
  PFA-En 0.85 & -20.27 & 6.37\% & 9.45\%\\
  \midrule
  PFA-KL & -2.19 & 35.02\% & 48.21\% \\
  \midrule
  \FG/A~\cite{Peng:ECCV:2018} & -0.04 & 32.78\% & 54.37\% \\
  \FG/B~\cite{Peng:ECCV:2018} & -0.30 & 23.37\% & 35.25\%\\
  \FG/C~\cite{Peng:ECCV:2018} & -0.44 & 16.30\% & 19.67\%\\
  \FG/D~\cite{Peng:ECCV:2018} & -1.22 & 14.23\% & 15.16\%\\
  \bottomrule  
  \end{tabular}
  \label{tab:resnet_imagenet_top5}
\end{table} 
\newpage

\pagebreak
\section{Architectures and Training Details}\label{app:training}
This appendix is meant to provide all details needed to reproduce the results presented in the paper.

\subsection{SimpleCNN}
\label{app:simplecnn}
The SimpleCNN network is used in~\Sec~\ref{sec:quantitative} in order to perform the experiments related to the upper-bound and the ablation studies.
   
SimpleCNN is composed of the following layers: 3 conv layers of size 96x3x3, a drop-out layer, 3 conv layers of size 192x3x3, drop-out layer, 1 conv layer of size 192x3x3, 1 conv layer of size 192x1x1, 1 conv layer of size [number of classes]x1x1, and finally an average pooling before the softmax layer. We use batchnorm and ReLU activations after every convolutional layer. 

In the following table we list the details for the training of the full and compressed architectures.

\begin{table}[H]
\centering
\begin{tabular}{lc}
			& \begin{tabular}{@{}c@{}} \bf{SimpleCNN on} \\ \bf{CIFAR-10 and CIFAR-100} \end{tabular} \\
			\hline
  Optimizer & \begin{tabular}{@{}c@{}} Nesterov \\ mom. 0.9, no decay \end{tabular} \\
  			\hline
  Learning rate & 0.1 \\
  			\hline
  \begin{tabular}{@{}c@{}}Learning rate \\ decay factor\end{tabular} & 0.1\\
    			\hline
  Epochs    & 50\\
    			\hline
  Epochs per decay & 30 \\
    			\hline
  Weights decay & 0.0001\\
    			\hline
  Batch size & 512\\
    			\hline
  Batch-norm & \begin{tabular}{@{}c@{}}moving average decay 0.99, \\ epsilon: 0.001\end{tabular} \\
    			\hline
  Drop-out & 0.5\\
    			\hline
  Augmentation & -- \\
  			\hline
\end{tabular}
\end{table}

\subsection{Training VGG-16 and ResNet-56 on CIFAR-10 and CIFAR-100}
\label{app:training_cifar}

In the following tables we list the details for the training of the full and compressed architectures used for the results presented in \Sec~\ref{sec:quantitative}, \Fig~\ref{fig:cifar10_100}.

\begin{table}[H]
\centering
\begin{tabular}{lc}
			& \begin{tabular}{@{}c@{}} \bf{VGG-16 on} \\ \bf{CIFAR-10 and CIFAR-100} \end{tabular} \\
			\hline
  Optimizer & \begin{tabular}{@{}c@{}} Nesterov \\ mom. 0.9, no decay \end{tabular} \\
  			\hline
  Learning rate & 0.1 \\
  			\hline
  \begin{tabular}{@{}c@{}}Learning rate \\ decay factor\end{tabular} & 0.1\\
    			\hline
  Epochs    & 160\\
    			\hline
  Epochs per decay & 90 \\
    			\hline
  Weights decay & 0.0001\\
    			\hline
  Batch size & 256\\
    			\hline
  Batch-norm & \begin{tabular}{@{}c@{}}moving average decay 0.99, \\ epsilon: 0.001\end{tabular} \\
    			\hline
  Drop-out & 0.5\\
    			\hline
  Augmentation & \begin{tabular}{@{}c@{}}
   we pad the image with 4 pixels around the boarder\\ and randomly crop a patch of size 32x32\\ and randomly flip the image 
   \end{tabular}\\
  			\hline
\end{tabular}
\end{table}

\begin{table}[H]
\centering
\begin{tabular}{lc}
			& \begin{tabular}{@{}c@{}} \bf{ResNet-56 on} \\ \bf{CIFAR-10 and CIFAR-100} \end{tabular} \\
			\hline
  Optimizer & \begin{tabular}{@{}c@{}} Nesterov \\ mom. 0.9, no decay \end{tabular} \\
  			\hline
  Learning rate & 0.1 \\
  			\hline
  \begin{tabular}{@{}c@{}}Learning rate \\ decay factor\end{tabular} & 0.1\\
    			\hline
  Epochs    & 160\\
    			\hline
  Epochs per decay & 90 \\
    			\hline
  Weights decay & 0.0005\\
    			\hline
  Batch size & 256\\
    			\hline
  Batch-norm & \begin{tabular}{@{}c@{}}moving average decay 0.99, \\ epsilon: 0.001\end{tabular} \\
    			\hline
  Drop-out & 0.5\\
    			\hline
  Augmentation & \begin{tabular}{@{}c@{}}
   we pad the image with 4 pixels around the boarder\\ and randomly crop a patch of size 32x32\\ and randomly flip the image 
   \end{tabular}\\
  			\hline
\end{tabular}
\end{table}

After compression it is possible that some ResNet blocks provide features maps with smaller number of channels than those forwarded by the respective skip-connections. In those cases, instead of padding the skip-connection we pad the output of the block before the combination with the skip-connection. This let us arbitrarily compress each block while ensuring the correct depth of the feature maps. 

\subsection{Training VGG-16 and ResNet-34 on ImageNet}
\label{app:training_imagenet}
In the following table we list the details for the training of the full and compressed architectures used for the results presented in \Sec~\ref{sec:quantitative}, \Fig~\ref{fig:imagenet_top1} and
in the App.~\ref{app:details}, \Fig~\ref{fig:top-5}.
For training VGG-16 with ImageNet we use distributed training~\cite{Ma:Corr:2018} with 8 machines and 8 GPUs each.

In the following tables we list the details for the training of the full and compressed architectures.

\begin{table}[H]
\centering
\begin{tabular}{lc}
			& \begin{tabular}{@{}c@{}} \bf{VGG-16 on} \\ \bf{ImageNet} \end{tabular} \\
			\hline
  Optimizer & Momentum \\
  			\hline
  Initial learning rate & 0.1 \\
  			\hline
  Max. learning rate & 1.6\\
  \hline
  \begin{tabular}{@{}c@{}}Learning rate \\ decay factor\end{tabular} & 0.9\\
    			\hline
  \begin{tabular}{@{}c@{}}Warming up \\ epochs\end{tabular} & 5\\
  \hline
  Epochs    & 90\\
    			\hline
  Epochs per decay & 2 \\
    			\hline
  Weights decay & 0.00004\\
    			\hline
  Batch size & 64\\
    			\hline
  Batch-norm & \begin{tabular}{@{}c@{}}moving average decay 0.99, \\ epsilon: 0.001\end{tabular} \\
    			\hline
  Drop-out & 0.5\\
    			\hline
  Augmentation & \begin{tabular}{@{}c@{}}
	We resize the shortest side of each image to 256\\ then we randomly crop an area of size 224x224\\ and randomly flip, finally we remove the average values\\ for each RGB channel: 123.68, 116.78, 103.94.
   \end{tabular}\\
  			\hline
\end{tabular}
\end{table}

\begin{table}[H]
\centering
\begin{tabular}{lc}
			& \begin{tabular}{@{}c@{}} \bf{ResNet-34-16 on} \\ \bf{ImageNet} \end{tabular} \\
			\hline
  Optimizer & Momentum \\
  			\hline
  Initial learning rate & 0.1 \\
  			\hline
  Max. learning rate & 1.6\\
  \hline
  \begin{tabular}{@{}c@{}}Learning rate \\ decay factor\end{tabular} & 0.85\\
    			\hline
  \begin{tabular}{@{}c@{}}Warming up \\ epochs\end{tabular} & 2\\
  \hline
  Epochs    & 180\\
    			\hline
  Epochs per decay & 4 \\
    			\hline
  Weights decay & 0.0001\\
  \hline
  Std weights in conv. & 0.1\\
    			\hline
  Batch size & 64\\
    			\hline
  Batch-norm & \begin{tabular}{@{}c@{}}moving average decay 0.99, \\ epsilon: 0.0001\end{tabular} \\
    			\hline
  Drop-out & 0.5\\
    			\hline
  Loss label smoothing & 0.1\\
  \hline
  Augmentation & \begin{tabular}{@{}c@{}}
	We resize the shortest side of each image to 256\\ then we randomly crop an area of size 224x224\\ and randomly flip, finally we remove the average values\\ for each RGB channel: 123.68, 116.78, 103.94.
   \end{tabular}\\
  			\hline
\end{tabular}
\end{table}

When using skip-connections with projections the application of PFA becomes easier than with padding. We analyze the output of the combination between the two branches (skip-connection and ResNet block) and reflect the compression back to the convolutions used for projection and that used as last step of the ResNet block.

\end{document}